\documentclass[11pt,a4paper]{article}
\usepackage{times,latexsym}
\usepackage{url}
\usepackage[T1]{fontenc}
\usepackage{amsmath,amssymb,amsfonts}
\usepackage{graphicx}
\usepackage{booktabs, enumitem}
\usepackage{multirow}
\usepackage{microtype}
\usepackage{url}
\usepackage{algorithm}
\usepackage[noend]{algpseudocode}
\usepackage{xcolor}
\usepackage{bm}
\usepackage{geometry}
\usepackage{rotating}
\usepackage[acceptedWithA]{tacl2021v1}

\newcommand{\softmax}{\operatorname{softmax}}
\newcommand{\adapt}{\textsc{ADAPT}}

\newcommand{\ft}{\textsc{SFT}}
\usepackage{tacl2021v1}

\usepackage{xspace,mfirstuc,tabulary}

\newif\iftaclinstructions
\taclinstructionsfalse 
\iftaclinstructions
\renewcommand{\confidential}{}
\renewcommand{\anonsubtext}{(No author info supplied here, for consistency with
TACL-submission anonymization requirements)}
\newcommand{\instr}
\fi

\title{\adapt: Learning Task Mixtures for Budget-Constrained Instruction Tuning}

\author{
\textbf{Pritam Kadasi\textsuperscript{1}\Thanks{Corresponding author.}} \quad
Abhishek Upperwal\textsuperscript{2} \quad
Mayank Singh\textsuperscript{1} \\
\textsuperscript{1}Lingo Research Group, Indian Institute of Technology Gandhinagar, India \\
\textsuperscript{2}Soket AI, India \\
\small{\texttt{\{pritam.k, singh.mayank\}@iitgn.ac.in}} \\
\small{\texttt{abhishek@soket.ai}}
}

\date{}

\begin{document}
\maketitle

\begin{abstract}

We propose \adapt{}, a meta-learning algorithm that \emph{learns} task sampling proportions under an explicit token budget for multi-task instruction tuning. Instead of fixing task weights by hand, \adapt{} maintains a continuous distribution over tasks and updates it via meta-gradients of a smooth worst-case validation objective, inducing an adaptive curriculum that allocates more tokens to useful tasks while avoiding collapse. We instantiate \adapt{} on three $\sim$1B-parameter open-weight LLMs (Gemma-3-1B, LLaMA-3.2-1B, Qwen-0.6B), training on 20 Natural Instructions task types under budgets of $1\%$, $5\%$, and $10\%$ of the available supervised tokens, and compare against strong supervised fine-tuning baselines with uniform and size-proportional mixing. We conduct evaluations on 11 out-of-domain benchmarks spanning reasoning, reading comprehension, code generation, and instruction following, we find that \adapt{} matches or slightly improves average downstream performance relative to the best static mixture, while using fewer effective training tokens and reallocating budget toward harder, benchmark-aligned tasks.


\end{abstract}
\section{Introduction}
\label{sec:intro}

Instruction finetuning has become a standard paradigm for aligning large language models (LLMs) with user-facing tasks, by fine-tuning on mixtures of supervised instruction--response pairs drawn from many datasets and task types \citet{NEURIPS2022_b1efde53,wang-etal-2023-self-instruct}. In practice, these mixtures are typically constructed using simple heuristics such as uniform sampling over datasets or sampling proportional to dataset size. While convenient, such heuristics ignore \emph{task utility}: some tasks may be redundant or too easy, whereas others are crucial for generalization to downstream evaluations.

This issue becomes especially acute in \emph{budget-constrained} settings. When practitioners have a fixed token budget (e.g., 1M tokens of instruction tuning) for small- to mid-sized open models, the choice of how to allocate tokens across tasks becomes a central design decision. In such regimes, we would like to answer: given many candidate tasks, which ones should we spend tokens on, and in what proportions, to maximize performance on challenging benchmarks such as MMLU-PRO~\cite{wang2024mmlupro}, BBH~\cite{suzgun-etal-2023-challenging}), GSM8K~\cite{cobbe2021trainingverifierssolvemath}.

In this work we propose \adapt{}, a meta-learning approach that \underline{adaptively} allocates a fixed token budget across tasks during multi-task instruction tuning. Instead of fixing task proportions by hand, \adapt{} maintains a learnable distribution over tasks and updates it using meta-gradients from validation loss, so that the mixture gradually shifts tokens toward tasks that are most beneficial for downstream generalization. \textsc{Adapt} earns its name by automatically increasing weight on harder, more informative tasks while down-weighting those that quickly saturate.

We instantiate \adapt{} for multi-task instruction tuning of three small open LLMs (Gemma-3-1B, LLaMA-3.2-1B, Qwen-0.6B) on 20 Natural Instructions task types, and compare it to strong supervised fine-tuning (\ft{}) baselines with uniform (SFT-U; equal tokens per task) and size-proportional (SFT-P; tokens proportional to training-set size) mixing, all trained under the \emph{same} token budgets of $1\%$, $5\%$, and $10\%$ of the total supervised training tokens. We then evaluate every model--budget--method combination on 11 out-of-domain LM-Eval benchmarks spanning general and mathematical reasoning, reading comprehension and QA, commonsense reasoning, code generation, and instruction following.

\paragraph{Empirical findings.}
Our experiments reveal three main trends:
\begin{itemize}[nolistsep]
    \item \textbf{Competitive downstream performance under the same budget.}
    Across three base models and budgets of $1$--$10\%$ of the training tokens, \adapt{} matches or slightly improves the macro-average score of the best static baseline (whichever of SFT-U or SFT-P is stronger at that budget), while winning or tying on a majority of tasks at moderate budgets (e.g., $9/11$ tasks vs.\ SFT-U and $10/11$ vs.\ SFT-P for \textsc{Gemma-1B} at $5\%$).
    \item \textbf{Substantial gains in training efficiency.}
    Tracking validation loss vs.\ tokens, \adapt{} reaches the best supervised validation loss using only $4$--$38\%$ of the tokens consumed by SFT, i.e., roughly $2.6$--$23\times$ fewer tokens depending on base model and budget, and area-under-curve metrics show that \adapt{} spends more of training in low-loss regimes.
    \item \textbf{Budget reallocation towards harder task types.}
    From the learned mixtures and group-wise scores, \adapt{} systematically shifts probability mass towards tasks aligned with challenging downstream benchmarks (e.g., reasoning and math), while keeping reading-comprehension and instruction-following competitive, indicating that the meta-gradient updates discover task trade-offs beyond static mixtures.
\end{itemize}

\paragraph{Contributions.}
In summary, this paper:
\begin{enumerate}[nolistsep]
    \item Formulates budget-constrained multi-task instruction tuning as a differentiable bilevel optimisation problem over task sampling proportions, and proposes \adapt{}, a one-step meta-gradient algorithm for learning continuous task mixtures under an explicit token budget.
    \item Provides an end-to-end, budget-aware implementation for small open-weight LLMs, including uniform and size-proportional supervised baselines within a shared training and evaluation pipeline, and releases code at~\url{https://github.com/pskadasi/ADAPT/}.
    \item Presents a detailed empirical study on three base models, 20 Natural Instructions task types, and 11 out-of-domain benchmarks, showing that \adapt{} achieves competitive or better downstream performance than strong static baselines while using substantially fewer tokens and reallocating budget towards harder task types.
\end{enumerate}

\section{Background}
\label{sec:background}
\subsection{Standard Multi-Task Instruction Tuning}

We consider $T$ supervised instruction-following tasks, indexed by $i \in \{1,\dots,T\}$.
Each task $i$ has a dataset
$D_i = D_i^{\text{train}} \cup D_i^{\text{val}} \cup D_i^{\text{test}}$,
where each example $(x,y)$ consists of an input instruction (and optional context) $x$ and a target response $y$.
We start from a pre-trained language model with parameters $\theta$ and fine-tune it with the standard next-token prediction loss on the concatenation of instruction and output.
Multi-task training is defined by a task-sampling distribution
$q = (q_1,\dots,q_T)$ with $q_i \ge 0$ and $\sum_i q_i = 1$: at each update, we sample a task index $i \sim q$, draw a minibatch from $D_i^{\text{train}}$, compute its loss, and update $\theta$ by gradient descent.

\subsection{Budget-Constrained Multi-Task Instruction Tuning}

For instruction tuning, where sequence lengths vary widely, it is natural to measure training cost in \emph{tokens} rather than steps.
We therefore fix a global token budget $B$ and track the total number of non-padding training tokens processed by the model; training stops once this budget is exhausted.

In our static supervised baselines, the task-sampling distribution $q$ is chosen a priori, typically as
$q_i = 1/T$ (uniform over tasks) or $q_i \propto |D_i^{\text{train}}|$ (proportional to train-set size).

The core question we study is whether we can \emph{learn} $q$ from validation feedback under a fixed budget $B$, so that the mixture automatically increases weight on harder, more informative tasks and down-weights those that quickly saturate.
This is precisely the role of \adapt{}, which replaces these static choices with a learned, adaptive sampling distribution.

\section{The \adapt{} Algorithm}
\label{sec:method}

Algorithm~\ref{alg:adapt} summarizes the meta-training loop under a global token budget $B$; below we detail its main components: the learned task mixture and inner update, the smooth worst-case validation objective, the entropy-regularised meta-update of the task logits, and the token-budgeted outer loop.

\subsection{Task Mixture and Inner Update}

Instead of fixing the task sampling distribution $q$ by hand, \adapt{} maintains a learnable vector of logits $w \in \mathbb{R}^T$ that defines a mixture over tasks via $p_i = \exp(w_i) / \sum_{j=1}^T \exp(w_j)$ for $i = 1,\dots,T$, so the task sampling distribution $p$ is learned during training rather than chosen manually. At each meta-iteration, we conceptually take one training minibatch $B_i^{\text{tr}}$ from each task $i \in \{1,\dots,T\}$ and compute per-task training losses $\ell_i(\theta) = L(\theta; B_i^{\text{tr}})$, where $L$ is the standard causal LM loss over instruction–response sequences. We then form the mixed training loss $L_{\text{mix}}(\theta, p) = \sum_{i=1}^{T} p_i \,\ell_i(\theta)$ and apply a differentiable inner update to the model parameters, $\theta' = \theta - \alpha \,\nabla_\theta L_{\text{mix}}(\theta, p)$ with inner learning rate $\alpha$, retaining the computation graph so that the dependence of $\theta'$ on the logits $w$ (via $p$) remains differentiable.

\subsection{Smooth Worst-Case Validation Objective}

After the inner update, we measure how good the update is for each task by evaluating the validation loss under $\theta'$, defining $v_i = L(\theta'; B_i^{\text{val}})$ for $i = 1,\dots,T$, where $B_i^{\text{val}}$ is a validation minibatch from $D_i^{\text{val}}$. To encourage robustness across tasks, we aggregate the per-task validation losses using a smooth approximation to the maximum, $J_\tau(v) = \tau \log \sum_{i=1}^{T} \exp(v_i / \tau)$, where $\tau > 0$ is a temperature hyperparameter; as $\tau \to 0$, $J_\tau(v)$ approaches $\max_i v_i$ (a hard worst-case objective), while larger $\tau$ interpolates toward an average over tasks.

\subsection{Entropy-Regularized Meta-Objective}

Directly minimizing a worst-case objective can cause the mixture to collapse onto a single task, especially early in training when validation losses are noisy, so we add an entropy regularizer over the task mixture, $H(p) = - \sum_{i=1}^{T} p_i \log p_i$. The overall meta-objective for updating the logits $w$ is then $L_{\text{meta}}(w) = J_\tau\big(v(w)\big) - \lambda\, H\big(p(w)\big)$, where $\lambda \ge 0$ controls the strength of entropy regularization. We update the logits by a gradient step $w \leftarrow w - \beta \,\nabla_w L_{\text{meta}}(w)$ with meta learning rate $\beta$, differentiating through both the inner update $\theta \mapsto \theta'$ and the softmax mapping $w \mapsto p(w)$.

\subsection{Token-Budgeted Training Loop}

 At a high level, each meta-iteration (i) forms a task mixture $p$ from the current logits $w$, (ii) performs a differentiable inner update on $\theta$ using the mixed training loss $L_{\text{mix}}$, (iii) evaluates validation losses $v_i$ under the updated parameters $\theta'$, (iv) updates the logits $w$ using the entropy-regularized smooth worst-case objective $L_{\text{meta}}$, and (v) optionally updates $\theta$ online and advances the global token counter, stopping when the budget $B$ is exhausted.

\begin{algorithm}[t]
\scriptsize
\caption{\adapt{}: one-step bilevel update with smooth validation objective}
\begin{algorithmic}[1]
\Require Tasks $\{\mathcal D_i^{\mathrm{tr}}, \mathcal D_i^{\mathrm{val}}\}_{i=1}^T$; token budget $B$
\Require Model parameters $\bm\theta$; task logits $\bm w$
\Require Inner LR $\alpha$; outer LR $\beta$
\Require Inner temperature $\tau>0$
\Require entropy weight $\lambda \ge 0$
\State \textbf{Initialize:} $\bm\theta \leftarrow \bm\theta_0$, $\bm w \leftarrow \bm 0$ \Comment{$p_i = 1/T$}
\State $tokens \gets 0$
\While{$tokens < B$}
  \State $\bm p \gets \softmax(\bm w)$
  \State Sample one train and val minibatch per task:
  \Statex \hspace{\algorithmicindent}$B_i^{\mathrm{tr}} \sim \mathcal D_i^{\mathrm{tr}},\;
  B_i^{\mathrm{val}} \sim \mathcal D_i^{\mathrm{val}}
  \quad \forall i \in \{1,\dots,T\}$
  \State $\ell_i(\bm\theta) \gets \mathcal L(\bm\theta; B_i^{\mathrm{tr}})$
  \State $L_{\mathrm{mix}}(\bm\theta,\bm p) \gets \sum_{i=1}^{T} p_i\,\ell_i(\bm\theta)$
  \State $\bm g \gets \nabla_{\bm\theta} L_{\mathrm{mix}}(\bm\theta,\bm p)$
  \State $\bm\theta' \gets \bm\theta - \alpha\,\bm g$
  \State $v_i \gets \mathcal L(\bm\theta'; B_i^{\mathrm{val}})$
  \State $J_\tau(\bm v) \gets \tau \log\!\big(\sum_{i=1}^T e^{v_i/\tau}\big)$
  \State $H(\bm p) \gets -\sum_{i=1}^T p_i \log p_i$
  \State $\mathcal L_{\mathrm{meta}} \gets J_\tau(\bm v) - \lambda\,H(\bm p)$
  \State $\bm w \leftarrow \bm w - \beta\,\nabla_{\bm w}\mathcal L_{\mathrm{meta}}$
  \State (Optional) $\bm\theta \leftarrow \bm\theta'$
  \State $tokens \gets tokens + \text{token\_count}\big(\{B_i^{\mathrm{tr}}\}_{i=1}^T\big)$
\EndWhile
\State \Return $\bm\theta$, $\bm p = \softmax(\bm w)$
\end{algorithmic}
\label{alg:adapt}
\end{algorithm}

\section{Experimental Setup}
\label{sec:experiments}

\subsection{Models}

We conduct experiments with three small open-weight decoder-only LLMs—Gemma-3-1B-PT, LLaMA-3.2-1B, and Qwen3-0.6B-Base—all in the $\sim$1B-parameter range. Each model is fine-tuned with left padding and a standard next-token prediction loss over the concatenation of instruction and response.

\subsection{Training Mixture and Baselines}

\paragraph{Datasets and task mixture.}

\begin{table}[tb]
\centering
\tiny
\begin{tabular}{lrrr}
\toprule
Category & \#Tasks & \#Instances & \#Tokens \\
\midrule
Translation                 & $394$ & $1{,}182{,}213$ & $72{,}549{,}385$ \\
Question Answering          & $207$ & $  470{,}108$   & $106{,}180{,}992$ \\
Program Execution           & $ 90$ & $  433{,}157$   & $35{,}066{,}354$  \\
Sentiment Analysis          & $ 75$ & $  253{,}432$   & $32{,}670{,}340$  \\
Question Generation         & $ 83$ & $  230{,}103$   & $56{,}131{,}362$  \\
Text Matching               & $ 43$ & $  173{,}171$   & $14{,}178{,}609$  \\
Text Categorization         & $ 46$ & $  154{,}556$   & $11{,}876{,}178$  \\
Commonsense Classification  & $ 24$ & $  130{,}524$   & $ 2{,}237{,}835$  \\
Toxic Language Detection    & $ 40$ & $  115{,}584$   & $ 5{,}148{,}102$  \\
Fill in The Blank           & $ 22$ & $   93{,}210$   & $13{,}687{,}063$  \\
Textual Entailment          & $ 27$ & $   92{,}651$   & $ 4{,}886{,}841$  \\
Information Extraction      & $ 34$ & $   91{,}850$   & $ 6{,}710{,}148$  \\
Text Completion             & $ 21$ & $   86{,}145$   & $ 8{,}631{,}760$  \\
Sentence Perturbation       & $ 15$ & $   80{,}789$   & $ 2{,}472{,}566$  \\
Title Generation            & $ 19$ & $   80{,}696$   & $20{,}869{,}481$  \\
Wrong Candidate Generation  & $ 27$ & $   73{,}546$   & $ 9{,}763{,}136$  \\
Sentence Composition        & $ 20$ & $   72{,}496$   & $ 5{,}066{,}324$  \\
Question Understanding      & $ 16$ & $   63{,}448$   & $ 3{,}239{,}390$  \\
Pos Tagging                 & $ 10$ & $   62{,}118$   & $ 2{,}583{,}225$  \\
Summarization               & $ 16$ & $   59{,}200$   & $43{,}445{,}932$  \\
\bottomrule
\end{tabular}
\caption{Dataset statistics by task category.}
\label{tab:dataset}
\end{table}

We use the \emph{Natural Instructions} dataset~\cite{wang-etal-2022-super}, which provides roughly $1{,}600$ tasks organized into $76$ task types, already curated in an instruction--response format. For this study, we focus on the $20$ largest task types by number of tasks (see Table~\ref{tab:dataset}).


\paragraph{Supervised fine-tuning baselines.}
We consider two standard static task-mixing strategies for supervised fine-tuning (SFT):
\begin{itemize}[nolistsep]
    \item \textbf{SFT-U (uniform)}: tasks are sampled uniformly, so that each of the $T=20$ training tasks receives approximately $B/T$ tokens.
    \item \textbf{SFT-P (size-proportional)}: tasks are sampled with probability proportional to their training-set size $|D_i^{\text{train}}|$, so that the expected number of tokens allocated to a task scales with its corpus size.
\end{itemize}
In both cases, sampling continues until the global budget $B$ is exhausted, at which point training stops.

\paragraph{ADAPT fine-tuning.}
For our method, \adapt{}, we also enforce the same global budget $B$ but let the allocation across tasks be \emph{learned}. \adapt{} maintains a vector of task logits and updates them via meta-gradients derived from a smooth worst-case validation objective, rather than fixing task proportions by hand. We refer to \adapt{}-trained models as \textbf{AFT} ($\adapt$ fine-tuning) in rest of the paper.

\subsection{Evaluation Benchmarks}

We evaluate pretrained base models and all fine-tuned variants (SFT-U, SFT-P, AFT) using the LM Evaluation ~\cite{biderman2024lessonstrenchesreproducibleevaluation} on a suite of 11 out-of-domain benchmarks grouped into six task types:
general knowledge \& reasoning (\textsc{GR}; MMLU-PRO~\cite{wang2024mmlupro}, BBH~\cite{suzgun-etal-2023-challenging}), mathematical reasoning (\textsc{MR}; GSM8K~\cite{cobbe2021trainingverifierssolvemath}, MATH~\cite{hendrycks2021measuring}), reading comprehension \& QA (\textsc{RQA}; SQuADv2~\cite{rajpurkar-etal-2018-know}, DROP~\cite{dua-etal-2019-drop}, TriviaQA~\cite{joshi-etal-2017-triviaqa}, NaturalQuestions~\cite{kwiatkowski-etal-2019-natural}), commonsense reasoning (\textsc{CR}; HellaSwag~\cite{zellers-etal-2019-hellaswag}), code generation (\textsc{CG}; HumanEval~\cite{chen2021evaluatinglargelanguagemodels}), and instruction following (\textsc{IF}; IFEval~\cite{zeng2024evaluating}).
For each benchmark, we report its standard metric (accuracy or exact match for \textsc{GR}, \textsc{MR}, \textsc{RQA}; Pass@1 for \textsc{CG}; accuracy-style metrics for \textsc{CR} and \textsc{IF}), together with a macro-average over all 11 benchmarks.

\subsection{Hyperparameters, Compute, and Training Details}

All models are trained with AdamW~\cite{loshchilov2018decoupled} and a cosine schedule with 200 warmup steps, decaying the learning rate to $0.1$ of its peak value over training. We use a global token budget $B \in {1,5,10}\%$ of the total supervised tokens in the 20-task mixture, with gradient clipping applied to both model parameters and task logits. Due to time and compute constraints, we only sweep the model and meta learning rates and keep all other hyperparameters fixed (Table~\ref{tab:hyperparams}). All runs use a single NVIDIA H100 GPU capped at 50,GB memory, reflecting the compute-constrained setting we target.

\begin{table}[tb]
\centering
\tiny
\begin{tabular}{lccc}
\toprule
\textbf{Hyperparameter} & \textbf{Symbol} & \textbf{AFT}  & \textbf{SFT} \\ 
\midrule

Model LR                & $\alpha$        & \begin{tabular}[c]{@{}c@{}}$\{3,5\}{\times}10^{-6}$,\\ $1{\times}10^{-5}$\end{tabular}      & \begin{tabular}[c]{@{}c@{}}$\{2,3,5,10,15\}$\\ $\times10^{-6}$\end{tabular} \\
Meta LR (logits)        & $\beta$         & \begin{tabular}[c]{@{}c@{}}$\{7.5, 5, 2.5\}{\times}10^{-3}$\end{tabular} & --                                                                        \\
Inner step LR           & $\gamma$        & $1{\times}10^{-4}$ (fixed)                                                                                     & --                                                                        \\
temperature             & $\tau$          & $0.3$ (fixed)                                                                                                  & --                                                                        \\
Entropy weight          & $\lambda$       & $1{\times}10^{-3}$ (fixed)                                                                                     & --                                                                        \\
Meta iterations         & --              & $20000$                                                                                                        & $20000$                                                                   \\
Tasks per step          & --              & $6$                                                                                                            & --                                                                        \\
Accumulation steps      & --              & $4$                                                                                                            & $8$                                                                       \\
Batch size              & --              & $1$                                                                                                            & $8$                                                                       \\
\bottomrule
\end{tabular}
\caption{Key hyperparameters and sweep ranges.}
\label{tab:hyperparams}
\end{table}

\section{Results}
\label{sec:results}

\subsection{Downstream Performance Under Budget Constraints}
\label{sec:downstream-budget}

\paragraph{Overall comparison to static baselines.}
\label{sec:rp1}
The zero-shot base models start from relatively modest averages over the 11 benchmarks:
$15.19$ for \textsc{Gemma-1B}, $15.16$ for \textsc{LLaMA-1B}, and $20.92$ for \textsc{Qwen-0.6B}.
Under budgeted instruction tuning, both AFT and static SFT baselines provide consistent gains.
At $1\%$ budget, AFT already improves the macro-average to $16.01$ (+0.83), $16.08$ (+0.92), and $21.36$ (+0.44) on \textsc{Gemma-1B}, \textsc{LLaMA-1B}, and \textsc{Qwen-0.6B}, respectively.
Across the three models and three budgets (9 settings), the best macro-average belongs to an SFT variant in 5/9 cases and to AFT in 4/9.
At $10\%$ on \textsc{LLaMA-1B}, SFT-P attains $16.46$ vs.\ $16.40$ for AFT (gap $0.06$); on \textsc{Gemma-1B}, SFT-P reaches $16.31$ vs.\ $16.18$ for AFT (gap $0.13$).
For \textsc{Qwen-0.6B} at $10\%$, AFT slightly \emph{outperforms} both SFT baselines, with $21.78$ vs.\ $21.54$ (SFT-U) and $21.66$ (SFT-P).
Overall, gaps between AFT and the best static mixture are small (typically $\leq 0.3$ points), and AFT consistently improves over zero-shot while closely tracking the strongest SFT curve.

\paragraph{Per-task win rates at fixed budget.}
\label{sec:rp2}
\begin{figure}[tb]
    \centering
    \includegraphics[width=0.8\linewidth]{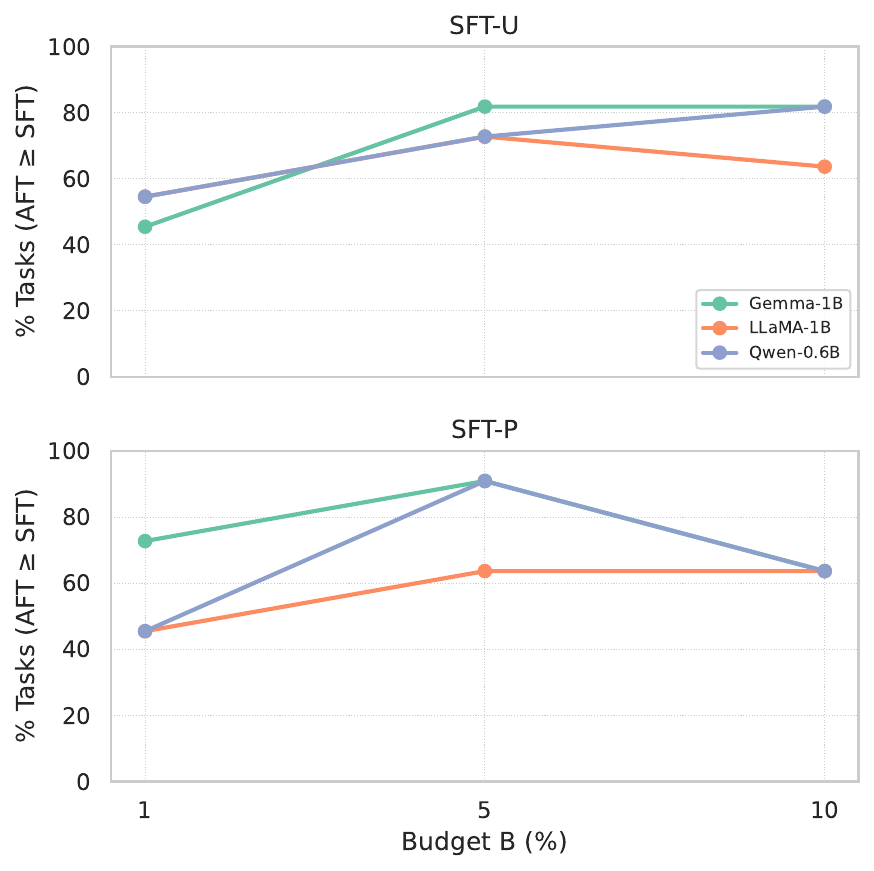}
    \caption{Win rates of AFT against static SFT baselines. For each base model and budget $b \in \{1,5,10\}\%$, we plot the fraction of the 11 benchmarks on which AFT matches or exceeds SFT-U and SFT-P at the same budget. AFT wins or ties on roughly half the tasks at $1\%$ and on a clear majority at $5$–$10\%$.}
    \label{fig:winrate}
\end{figure}
Average scores hide how gains are distributed across benchmarks, so we also compute, for each base and budget, the fraction of tasks (out of 11) on which AFT matches or exceeds each SFT baseline at the \emph{same} budget (Figure~\ref{fig:winrate}).

Against SFT-U at $5\%$, AFT matches or beats the baseline on
$9/11$ tasks for \textsc{Gemma-1B} (81.8\%), $8/11$ for \textsc{LLaMA-1B} (72.7\%), and
$8/11$ for \textsc{Qwen-0.6B} (72.7\%).
At $10\%$, this remains high: $9/11$ for \textsc{Gemma-1B} and \textsc{Qwen-0.6B} (81.8\%), and $7/11$ for \textsc{LLaMA-1B} (63.6\%).
Even at $1\%$, AFT wins or ties SFT-U on about 5--6 of 11 tasks.

Relative to SFT-P, at $5\%$ AFT matches or exceeds the baseline on $10/11$ tasks for both \textsc{Gemma-1B} and \textsc{Qwen-0.6B} (90.9\%), and on $7/11$ tasks (63.6\%) for \textsc{LLaMA-1B}.
At $10\%$, AFT still wins or ties on $7/11$ tasks across all three bases.
Thus, while static SFT can have a slight edge in macro-average, AFT yields a more balanced profile, matching or surpassing the static mixture on most individual benchmarks at the same budget.

\paragraph{Gains over zero-shot.}
\label{sec:rp3}
Combining averages and win rates, moving from zero-shot to any budgeted tuning regime (AFT or SFT) reliably improves performance by roughly $0.5$--$2$ macro-average points, with the largest gains for \textsc{LLaMA-1B} and \textsc{Qwen-0.6B}.
AFT achieves these gains while remaining competitive with, and in many settings exceeding, the best static mixture on a majority of tasks.

\begin{figure*}[tb]
    \centering
    \includegraphics[width=0.8\textwidth]{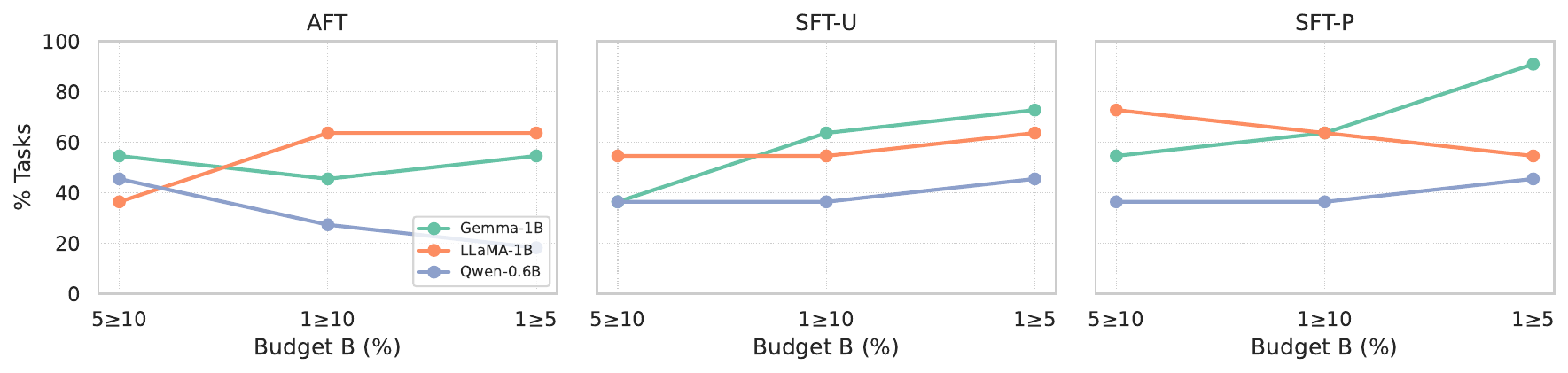}
    \caption{Budget efficiency across models and methods. For each base model and budget pair $(1\%\!,5\%)$, $(1\%\!,10\%)$, and $(5\%\!,10\%)$, bars show the percentage of tasks on which the smaller-budget run matches or exceeds the larger-budget run, highlighting diminishing returns beyond about $5\%$ of the training tokens.}
    \label{fig:bud_eff}
\end{figure*}

\paragraph{Budget sensitivity and diminishing returns.}
\label{sec:rp4}
To assess budget sensitivity for AFT, we compare, for each base and each pair of budgets $(b_{\text{small}},b_{\text{large}})\in\{(1,5),(1,10),(5,10)\}$, the fraction of tasks where the smaller-budget run matches or exceeds the larger-budget run (Figure~\ref{fig:bud_eff}).
On \textsc{Gemma-1B}, the $5\%$ AFT model matches or outperforms the $10\%$ model on $6/11$ tasks (54.6\%), and the $1\%$ model matches or beats the $5\%$ and $10\%$ variants on $6/11$ and $5/11$ tasks, respectively.
\textsc{LLaMA-1B} shows a similar pattern between $1\%$ and $5\%$ (7/11 tasks where $1\% \ge 5\%$), with clearer gains when moving from $5\%$ to $10\%$ on the remaining tasks.
For \textsc{Qwen-0.6B}, higher budgets help more, but even there the $5\%$ run matches or exceeds the $10\%$ run on 45.5\% of tasks (5/11).

Static SFT baselines also exhibit diminishing returns: for both SFT-U and SFT-P, the $5\%$ models already match or surpass the $10\%$ models on roughly one-third to two-thirds of the tasks (about 4--7 out of 11, depending on the base), so only a minority of benchmarks clearly benefit from increasing the budget from $5\%$ to $10\%$.
Taken together with the AFT results, this suggests that most downstream gains are realized by around $5\%$ of the training tokens, with $10\%$ offering only modest additional improvements.

\begin{figure*}[tb]
    \centering
    \includegraphics[width=0.8\textwidth]{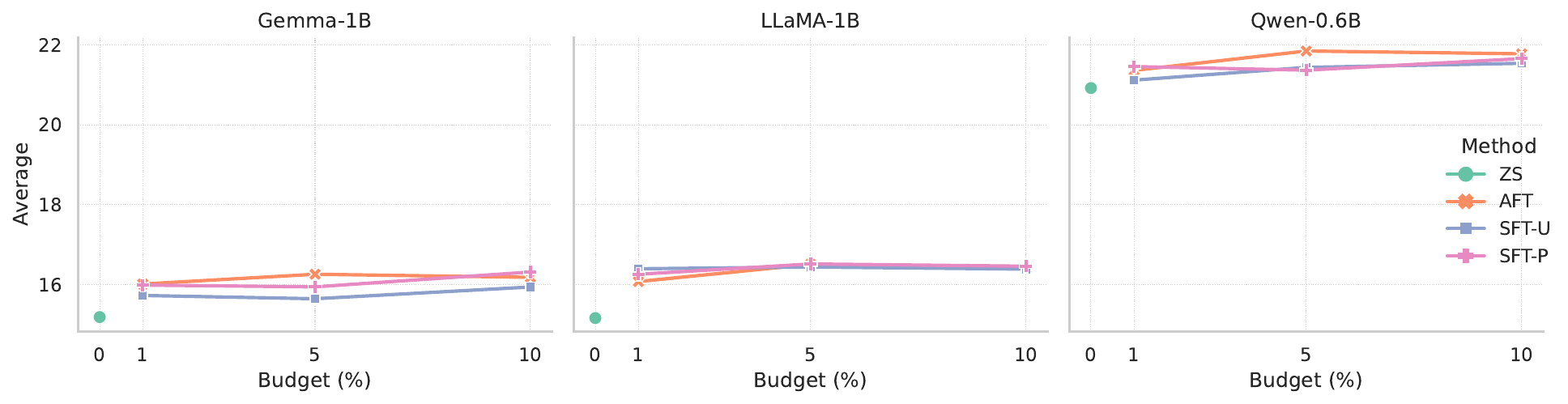}
    \caption{Average evaluation score as a function of budget for AFT, SFT-U, and SFT-P on each base model. Curves are shown for budgets $\{0,1,5,10\}\%$ of the supervised tokens (with $0\%$ corresponding to zero-shot), illustrating steep gains from $0\%\!\to\!5\%$ and saturation between $5\%$ and $10\%$ while AFT closely tracks the best static baseline.}

    \label{fig:avg_scores}
\end{figure*}

\paragraph{Budget--performance curves.}
\label{sec:rp5}
Figure~\ref{fig:avg_scores} summarizes these findings by plotting average LM-Eval score vs.\ budget $\{0,1,5,10\}\%$ for AFT, SFT-U, and SFT-P (zero-shot at $0\%$).
Across all three bases, the curves rise steeply from $0\%$ to $5\%$ and then saturate between $5\%$ and $10\%$.
AFT closely tracks the best static baseline at each budget, and often dominates it in the low- and mid-budget regime when measured by per-task win rate.
Overall, AFT matches or surpasses strong static SFT baselines under the same token budget, especially in the $1$--$5\%$ range where most gains over zero-shot are achieved.

\subsection{Training Efficiency}
\label{sec:training-efficiency}

Beyond final downstream performance, we evaluate how efficiently each method uses its training budget.
For each run we log the validation loss as a function of the cumulative number of training tokens
and analyze the resulting loss--vs--budget curves.

\begin{figure*}[tb]

    \centering
    \includegraphics[width=0.8\textwidth]{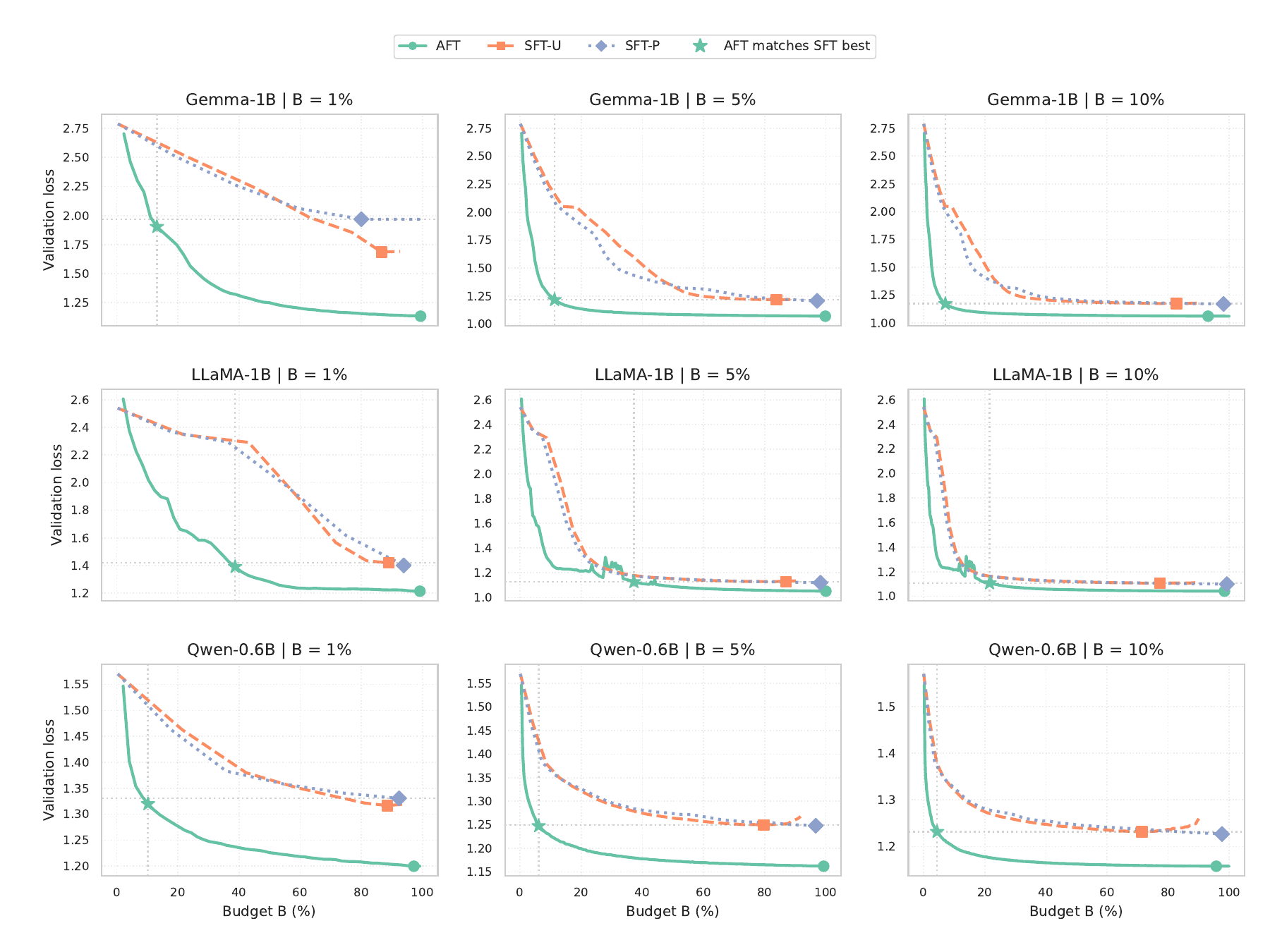}
    \caption{Validation loss as a function of cumulative training tokens for AFT, SFT-U, and SFT-P across base models and budgets. AFT consistently descends faster and reaches low-loss regimes with fewer tokens than the static SFT baselines, indicating better convergence efficiency under the same token budget.}

    \label{fig:eff_curves}
\end{figure*}

Figure~\ref{fig:eff_curves} shows a 3$\times$3 grid of training curves:
rows correspond to base models (\textsc{Gemma-1B}, \textsc{LLaMA-1B}, \textsc{Qwen-0.6B}) and
columns to budget slices (1\%, 5\%, 10\% of the full corpus).
Within each panel, we plot AFT, SFT-U, and SFT-P.
Visually, AFT consistently descends faster and flattens earlier: it reaches low-loss regimes
after a small fraction of the budget, while the supervised baselines continue to train for the
full budget with comparatively modest additional gains.
We quantify this behavior using three complementary metrics derived from the same curves:
(i) the fraction of tokens required for AFT to match the best supervised validation loss,
(ii) the area under the loss--vs--tokens curve (AUC), and (iii) the validation loss of AFT
at mid-budget (50\% of tokens).

\begin{figure*}[tb]
    \centering
    \includegraphics[width=0.8\textwidth]{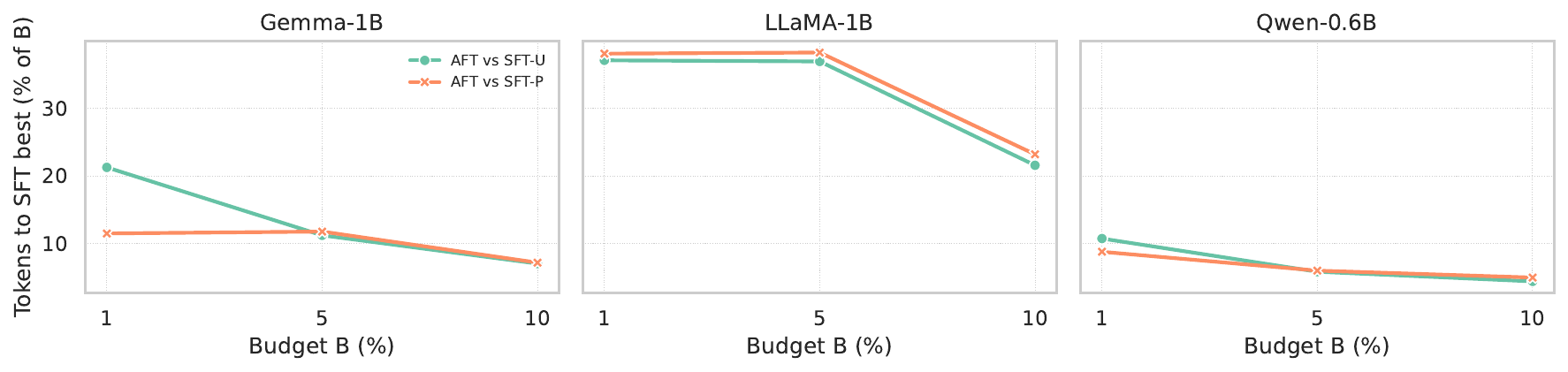}
    \caption{Tokens (as a percentage of the budget $B$) required by AFT to match the best validation loss of the strongest SFT baseline (SFT-U or SFT-P) across budgets $B \in \{1,5,10\}\%$ for each base model. Lower values indicate that AFT reaches supervised quality with fewer tokens.}

    \label{fig:tok_thr}
\end{figure*}

\paragraph{Tokens to match supervised quality.}
\label{sec:rp6}
For every base model and budget (1\%, 5\%, 10\%), we compute the earliest point at which AFT's
validation loss falls below the best loss achieved by the corresponding supervised run
(SFT-U or SFT-P).
On \textsc{Gemma-1B}, AFT reaches the best supervised loss after consuming only
$\approx$7--11\% of the tokens, corresponding to roughly $8$--$14\times$ fewer tokens than running
the full supervised schedule.
On \textsc{Qwen-0.6B} the effect is even stronger: AFT matches the best supervised loss after
only $\approx$4--11\% of the tokens (up to $\sim$23$\times$ reduction), while on \textsc{LLaMA-1B}
it does so after $\approx$22--38\% of the budget ($\sim$2.6--4.6$\times$ reduction).
Figure~\ref{fig:tok_thr} visualizes these gains across models and budgets.

\begin{figure*}[tb]
    \centering
    \includegraphics[width=0.8\textwidth]{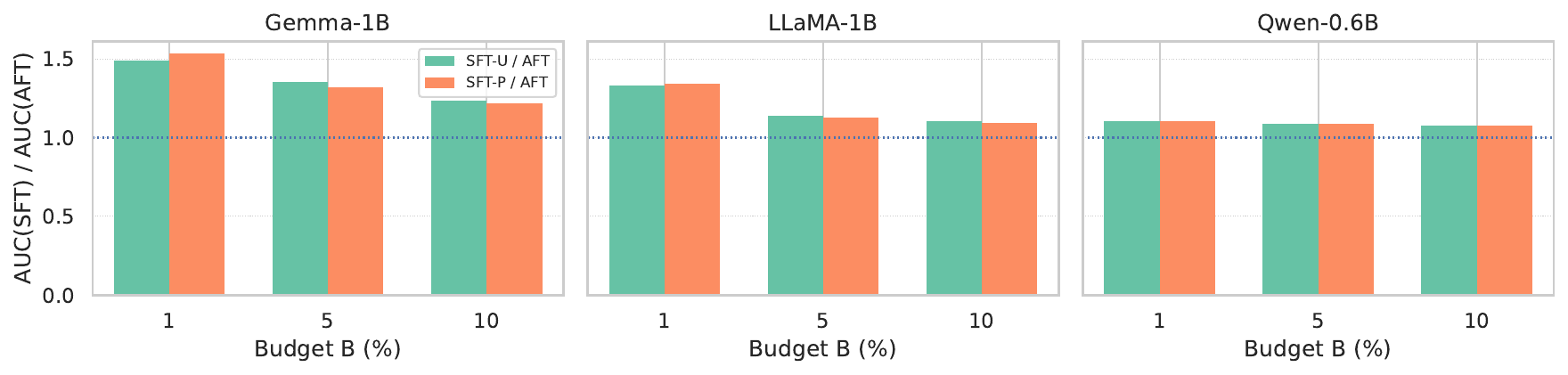}
    \caption{AUC efficiency ratios for validation loss, shown as $\mathrm{AUC}(\text{SFT}) / \mathrm{AUC}(\text{AFT})$ across budgets and base models. Bars above $1$ indicate that AFT spends a larger fraction of training at low validation loss than the corresponding SFT baseline (i.e., is more training-efficient).}
    \label{fig:auc_eff}
\end{figure*}

\paragraph{AUC: overall convergence efficiency.}
\label{sec:rp7}
We summarize the entire training trajectory using the area under the loss--vs--tokens curve
(up to the full budget).
Lower AUC implies that the model spends more of its training time at low validation loss, i.e.,
it converges faster and wastes fewer updates.
Across all three base models and all budget settings, AFT consistently achieves a lower AUC than
both SFT-U and SFT-P.
Expressed as a ratio $\mathrm{AUC}(\text{SFT}) / \mathrm{AUC}(\text{AFT})$
(Figure~\ref{fig:auc_eff}, supervised baselines are typically about
1.1--1.5$\times$ less efficient, corresponding to an AUC reduction of roughly 8--35\% in favor of AFT.
This matches the qualitative picture in Figure~\ref{fig:eff_curves}, where Adapt spends a larger
fraction of training in the low-loss region.

\paragraph{Mid-budget performance.}
\label{sec:rp8}
Finally, we compare the validation loss of AFT at 50\% of the token budget with the \emph{best}
validation loss reached by each supervised run over its entire training trajectory.
For all three base models and all budget slices, AFT at mid-budget already achieves a lower loss
than the final SFT-U and SFT-P checkpoints.
This indicates that a substantial fraction of the supervised updates (often the second half of the run)
provides little to no benefit, whereas AFT front-loads useful updates and converges earlier.

Taken together with the downstream evaluation in \S\ref{sec:downstream-budget}, these results show that AFT attains comparable or better accuracy than SFT while using substantially fewer training tokens and converging faster, making it an attractive choice under tight compute or data budgets.

\subsection{Where Does AFT Help? -- Task-Type and Difficulty Analysis}
\label{sec:task-type-analysis}

\begin{figure}[tb]
    \centering
    \includegraphics[width=0.8\linewidth]{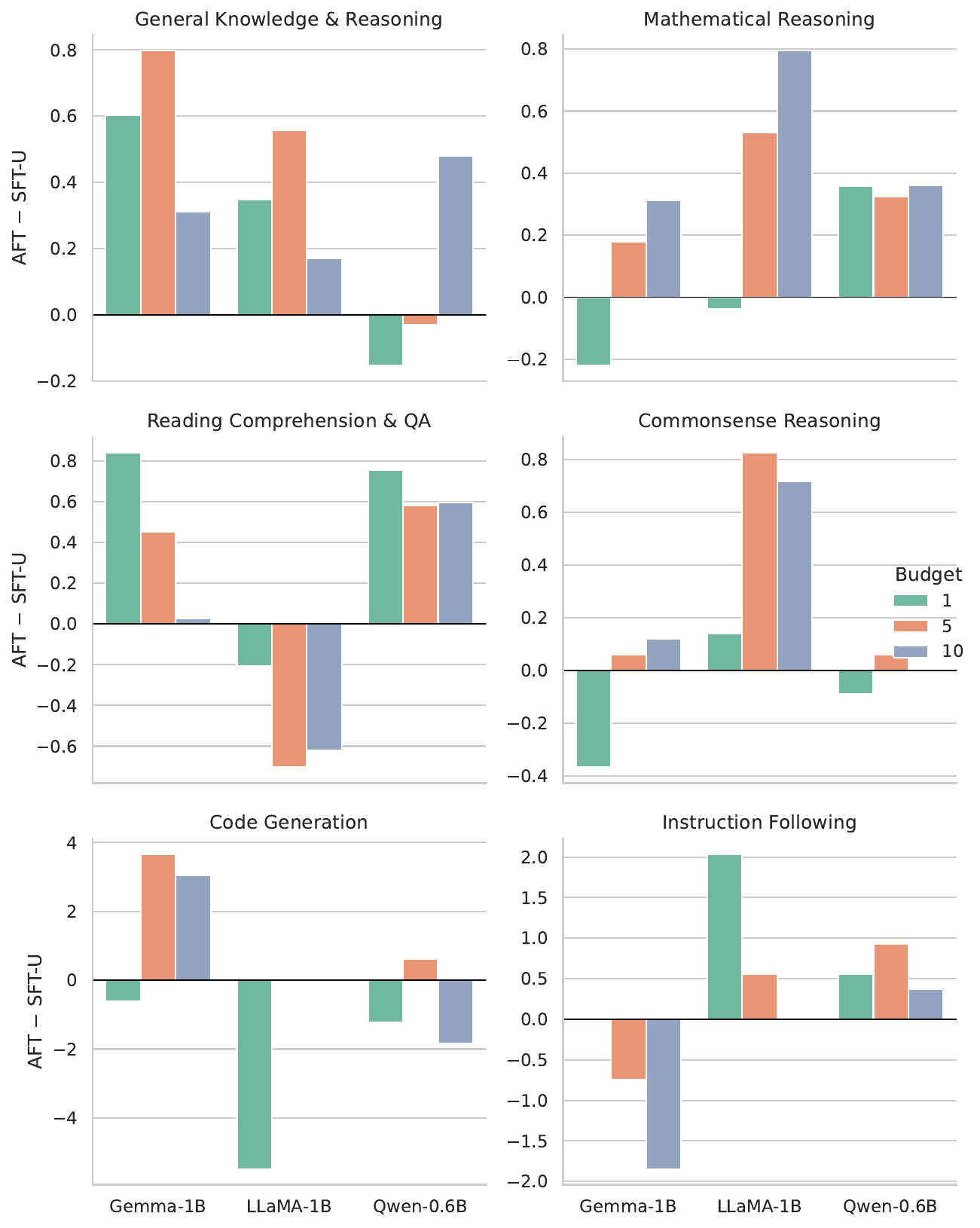}
    \caption{Task-type and difficulty analysis relative to uniform SFT (SFT-U). For each base model on the $x$-axis, we plot the group-wise score difference $\text{AFT} - \text{SFT-U}$ in percentage points (pp). Positive values indicate that ADAPT outperforms uniform mixing for that task type at the corresponding budget.}
    \label{fig:group_u}
\end{figure}

\begin{figure}[tb]
    \centering
    \includegraphics[width=0.8\linewidth]{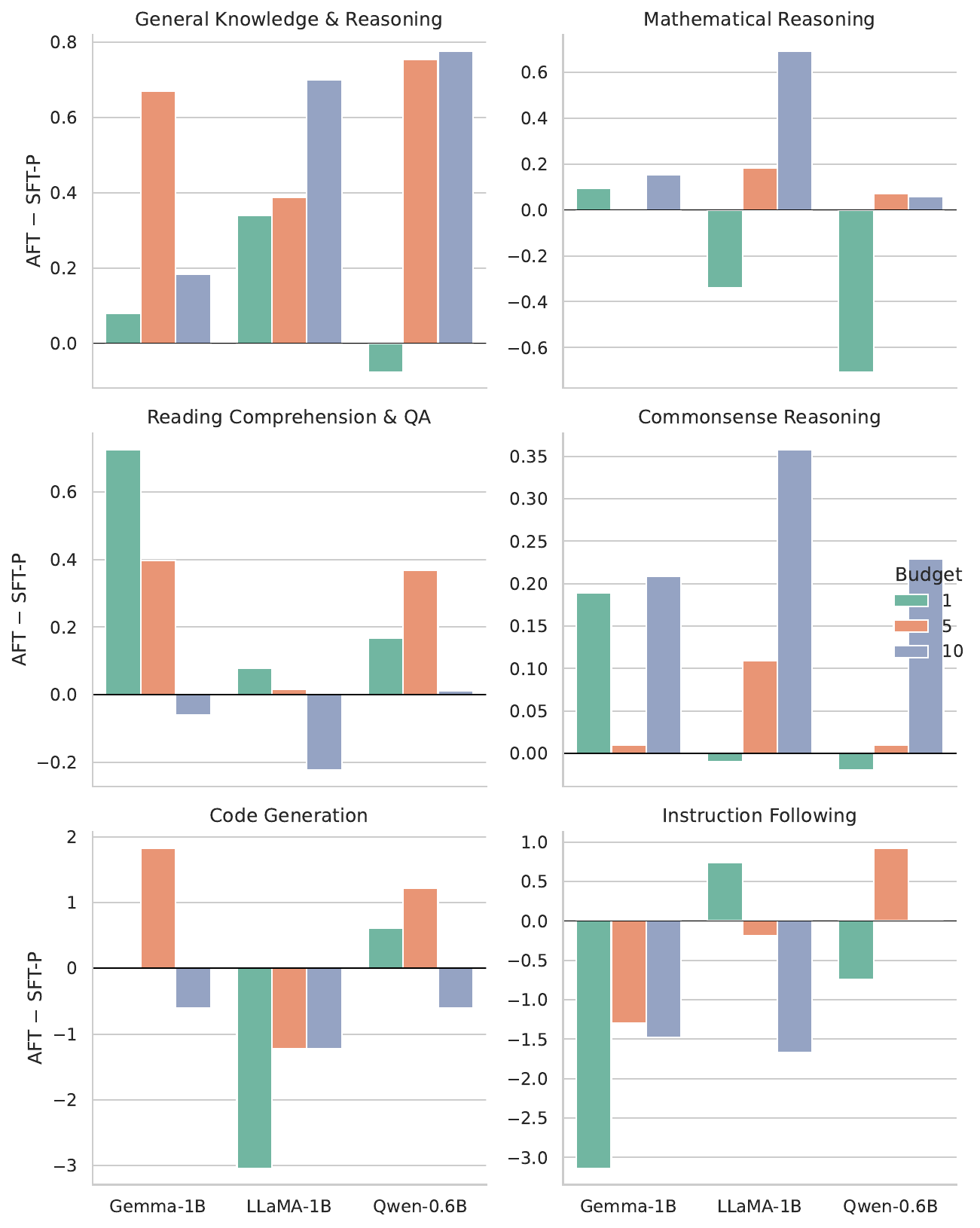}
    \caption{Task-type and difficulty analysis relative to size-proportional SFT (SFT-P). For each base model on the $x$-axis, we plot the group-wise score difference $\text{AFT} - \text{SFT-P}$ in percentage points (pp). Positive values indicate that ADAPT outperforms size-proportional mixing for that task type at the corresponding budget.}
    \label{fig:group_p}
\end{figure}

To understand \emph{where} AFT helps beyond overall averages, for each base model, budget, and task group we compute the average score difference between AFT and the corresponding SFT baseline (\textsc{U}/\textsc{P}). Figures~\ref{fig:group_u} and~\ref{fig:group_p} show these group-wise deltas.

\paragraph{Comparison to uniform SFT.}
\label{sec:rp9}
Relative to SFT-U, AFT gives its most consistent gains on \textsc{GR} and \textsc{RQA}. For \textsc{Gemma-1B}, AFT at $5\%$ is ahead by roughly $\sim\!0.8$ points on \textsc{GR} and $\sim\!0.45$ on \textsc{RQA}, with smaller but still positive gains at $1\%$ and $10\%$. For \textsc{Qwen-0.6B}, AFT is slightly behind on \textsc{GR} at $1\%$ but becomes positive by $5$--$10\%$, while consistently improving \textsc{RQA} across all budgets. On \textsc{LLaMA-1B}, AFT again improves \textsc{GR} and, especially at higher budgets, \textsc{MR}, while trading off at most about $\pm 1$ point on \textsc{RQA}. Overall, AFT reallocates budget toward reasoning-heavy groups while keeping \textsc{RQA} competitive.

For \textsc{CG}, AFT yields large gains for \textsc{Gemma-1B} at $5$--$10\%$ (roughly $+3$ points on HumanEval), and smaller positive effects for \textsc{Qwen-0.6B} at $5\%$. On \textsc{LLaMA-1B}, AFT underperforms SFT-U on HumanEval at $1\%$, suggesting that aggressively reallocating tokens away from code can hurt very small code subsets. For \textsc{IF}, AFT often improves over SFT-U for \textsc{LLaMA-1B} and \textsc{Qwen-0.6B} (gains of $\sim\!0.5$--$2$ points, depending on budget), while \textsc{Gemma-1B} shows mild regressions at higher budgets, consistent with the per-task trends in~\S\ref{sec:downstream-budget}.


\paragraph{Comparison to proportional SFT.}
\label{sec:rp10}
Against SFT-P, AFT shows a similar pattern but with fewer clear losses. Across all three bases, AFT is consistently neutral or better on \textsc{GR} and typically non-negative on \textsc{MR}, with noticeable gains at moderate budgets (e.g., \textsc{LLaMA-1B} and \textsc{Qwen-0.6B} at $5$--$10\%$). On \textsc{RQA}, AFT usually matches or slightly surpasses SFT-P: for \textsc{Gemma-1B} and \textsc{Qwen-0.6B}, group-wise differences are mostly positive at $1$--$5\%$, and only slightly negative in a few high-budget settings. In \textsc{CG}, AFT again improves HumanEval for \textsc{Gemma-1B} and \textsc{Qwen-0.6B} at $5\%$ (roughly $+1.8$ and $+1.2$ points, respectively), while \textsc{LLaMA-1B} remains better served by proportional mixing for this single task. \textsc{IF} scores remain mixed: AFT is sometimes ahead (e.g., \textsc{LLaMA-1B} at $1\%$, \textsc{Qwen-0.6B} at $5\%$), sometimes slightly behind, indicating that \textsc{IF} is sensitive to how budget is redistributed away from broad instruction-like tasks.

\paragraph{Task type and difficulty.}
\label{sec:rp11}
Overall, these group-wise comparisons indicate that AFT's largest and most reliable gains arise in the more reasoning-heavy categories (\textsc{GR}, \textsc{MR}, and, for some bases, \textsc{CG}), while maintaining competitive performance on \textsc{RQA}. Under a fixed token budget, AFT thus tends to shift the mixture toward tasks that are relatively harder for the base models, yielding broader improvements in reasoning and problem-solving ability without materially sacrificing performance on easier or already higher-scoring groups.

\subsection{Ablation on Entropy Regularization}

\begin{figure}[tb]
    \centering
    \includegraphics[width=0.8\linewidth]{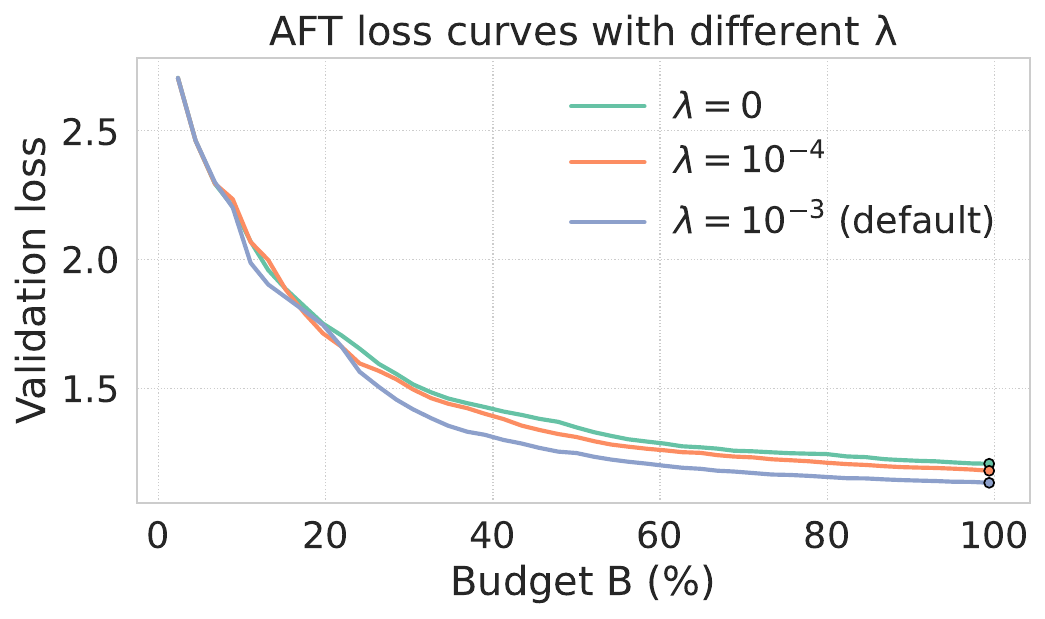}
    \caption{Validation loss vs.\ cumulative training tokens for different entropy weights $\lambda$ on \textsc{Gemma-1B} at a $1\%$ budget. The default setting $\lambda = 10^{-3}$ converges fastest and to the lowest loss, $\lambda = 10^{-4}$ is intermediate, and $\lambda = 0$ performs worst.}
    \label{fig:lambda_loss}
\end{figure}

\begin{figure*}[tb]
    \centering
    \includegraphics[width=0.8\linewidth]{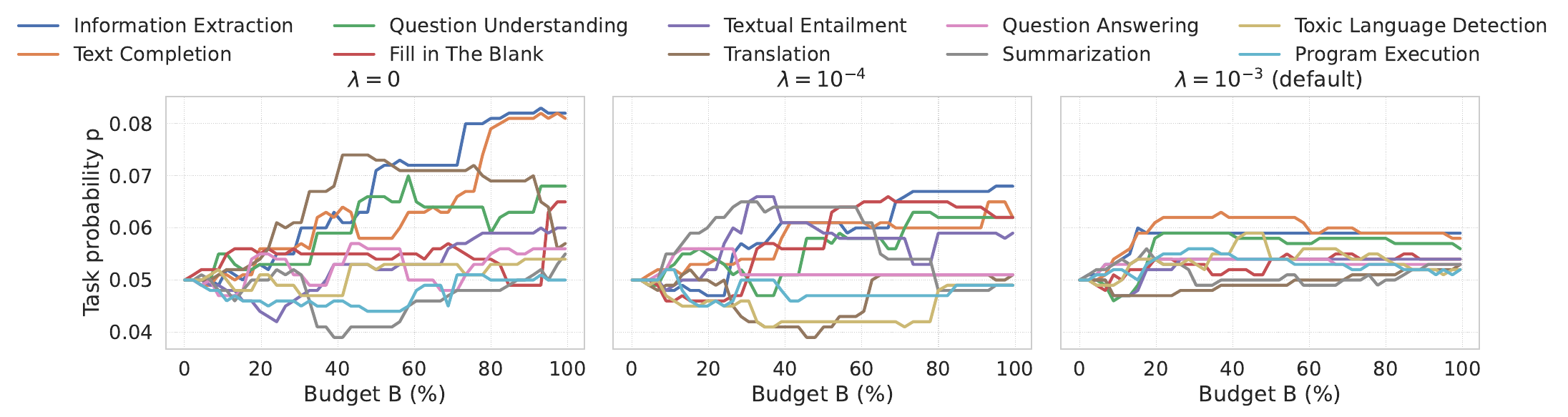}
    \caption{Evolution of task weights $p_i$ over training for \textsc{Gemma-1B} at a $1\%$ budget, showing the top-$10$ tasks under different entropy weights $\lambda$. Larger $\lambda$ yields more diverse mixtures, while $\lambda = 0$ concentrates mass on fewer tasks.}
    \label{fig:lambda}
\end{figure*}

Our sampler includes an entropy penalty $\lambda H(p)$ on the task mixture $p$ to discourage collapse onto a few tasks. We run a focused ablation on \textsc{Gemma-1B} at a 1\% token budget, comparing three settings:
(i) no entropy ($\lambda = 0$),
(ii) a very small entropy weight ($\lambda = 10^{-4}$), and
(iii) our default setting ($\lambda = 10^{-3}$).
All other hyperparameters are fixed.
For each run we track validation loss vs.\ budget and summarize the learned mixtures via the effective number of tasks $N_{\text{eff}} = 1 / \sum_i p_i^2$, which equals $T$ for a uniform mixture and approaches $1$ as mass collapses onto a single task.
Table~\ref{tab:lambda_sweep} and Figures~\ref{fig:lambda_loss},~\ref{fig:lambda} report these results.

\begin{table}[t]
  \centering
  \scriptsize
  \begin{tabular}{@{}cccc@{}}
    \toprule
    $\lambda$ & $N_{\text{eff}}$ & Tokens used & Entropy \\
    \midrule
    $0$                  & $18.35$ & $99.4\%$ & $2.950$ \\
    $10^{-4}$            & $19.52$ & $99.4\%$ & $2.983$ \\
    $10^{-3}$\,(default) & $19.70$ & $99.4\%$ & $2.990$ \\
    \bottomrule
  \end{tabular}
  \caption{Effect of entropy weight $\lambda$ on the learned task mixture for \textsc{Gemma-1B} at a $1\%$ budget. Higher $N_{\text{eff}}$ means the mixture covers more tasks, while lower $N_{\text{eff}}$ (approaching $1$) indicates collapse onto a single task.}
  \label{tab:lambda_sweep}
\end{table}

At 1\% budget, the default $\lambda=10^{-3}$ yields the most diverse yet clearly non-uniform mixture ($N_{\text{eff}} = 19.70$, $H(p) = 2.990$) while using 99.4\% of the available tokens. Reducing the weight to $\lambda = 10^{-4}$ slightly reduces diversity ($N_{\text{eff}} = 19.52$, $H(p) = 2.983$), and removing entropy ($\lambda=0$) produces a more concentrated mixture ($N_{\text{eff}} = 18.35$, $H(p) = 2.950$) despite the same token usage. The top-$k$ trajectories (10 highest-probability tasks) make this visible: with $\lambda=0$ probability mass concentrates on a few high-loss tasks and several others go near zero, whereas $\lambda=10^{-3}$ maintains substantial mass on a broader set of tasks throughout training (Figure~\ref{fig:lambda}).

The validation-loss curves (Figure~\ref{fig:lambda_loss}) show a consistent ordering: $\lambda=10^{-3}$ achieves the lowest loss, $\lambda=10^{-4}$ is slightly worse, and $\lambda=0$ performs worst across the 1\% run. This ablation supports our design choice: even for a single base model and budget slice, a moderate entropy weight ($\lambda = 10^{-3}$) is needed to avoid brittle mode collapse while still allowing the meta-gradient to learn useful, non-uniform task mixtures. We therefore use $\lambda = 10^{-3}$ in all main experiments.


\section{Related Work}
\label{sec:related}

\paragraph{Instruction tuning and multi-task learning.}
Supervised instruction tuning has become a standard recipe for aligning LLMs with human intent, by fine-tuning on large mixtures of instruction--response pairs drawn from diverse sources~\citep{wei2022finetuned,sanh2022multitaskpromptedtrainingenables,NEURIPS2022_b1efde53,wang-etal-2022-super,chung2022scalinginstructionfinetunedlanguagemodels,iyer2023optimlscalinglanguagemodel}. These collections aggregate tasks spanning QA, reasoning, code, and dialogue, and are often reused or extended in open suites such as Tulu, Alpaca, Orca, and related variants~\citep{wang2023how,taori2023stanford,mukherjee2023orcaprogressivelearningcomplex,zhang2025the}. A central challenge is managing heterogeneity and imbalance: tasks differ widely in size, difficulty, and quality, yet are typically combined using simple heuristics (uniform or size-proportional sampling). Our work follows this multi-task instruction-tuning paradigm, but instead of fixing task proportions by hand, \adapt{} learns task mixtures under an explicit token budget.

\paragraph{Data mixture optimisation and bilevel/meta-learning.}
A growing line of work optimises data \emph{mixtures} across domains or tasks rather than treating them as fixed. In pre-training, DoReMi~\citep{xie2023doremi} learns domain weights with a small proxy model trained via group DRO and transfers them to larger models; CLIMB~\citep{diao2025climbclusteringbasediterativedata} and related regression-based predictors refine cluster-level mixtures using proxy evaluations; and online or bandit-style schemes adapt sampling during training~\citep{albalak2023efficient,li2025pike,fan2025grape}. For supervised fine-tuning, Data Mixing Optimization for SFT~\citep{li2025datamixingoptimizationsupervised} models validation loss as a function of dataset weights and scale, IDEAL~\citep{ming2025idealdataequilibriumadaptation} and VersaTune~\citep{lu-etal-2025-versatune} iteratively adjust domain-level mixtures based on performance and forgetting, and Dynamic Data Mixing for MoE models~\citep{zhu-etal-2025-dynamic} uses routing statistics (gate loads) to set sampling weights. More broadly, bilevel and meta-learning methods such as ScaleBiO~\citep{pan-etal-2025-scalebio}, MOS~\citep{wu2024mixtureofskillslearningoptimizedata}, and HBO~\citep{wang2025hbohierarchicalbalancingoptimization} reweight data or learn sampling policies via gradient-based or RL-based hyperparameter optimisation. Compared to these methods, \adapt{} uses a simple one-step differentiable bilevel update over task logits, tied to a smooth worst-case validation objective under strict token budgets for small open-weight LLMs.

\section{Limitations}
\label{sec:limitations}

Our study has several limitations. First, we only experiment with small open-weight models ($\sim$0.6--1B parameters) on a single 20-task Natural Instructions mixture, evaluated on 11 English LM-Eval benchmarks, so it is unclear how \adapt{} scales to larger models, multilingual or multimodal settings, or more heterogeneous web-scale mixtures. Second, \adapt{} relies on per-task validation splits to drive the meta-objective; in low-resource regimes where such validation data are scarce or noisy, the method may be harder to apply, and we do not explore proxy or unsupervised alternatives. Third, we use a specific design (one-step inner update, smooth worst-case validation loss, entropy regularization) and tune only model and meta learning rates under a fixed hardware setup (single H100 GPU, 50,GB cap), so a broader search over objectives, inner-loop depth, and training hyperparameters could change the relative ranking between \adapt{} and the supervised baselines. Finally, our evaluation focuses on accuracy-style benchmark metrics; we do not study calibration, robustness, safety, or fairness, nor combine \adapt{} with instance-level data selection methods, which we leave for future work.

\section{Ethical Considerations}
\label{sec:ethics}

Our work reallocates tokens across tasks for instruction tuning of open LLMs and introduces no new data or architectures, but any such model inherits biases and safety issues from its pretraining corpus and task mixture. Care is needed to ensure that high-utility tasks do not overemphasize sensitive or harmful content, so we recommend pairing \adapt{} with standard safety filtering and human oversight in deployment.

\section{Conclusion}
\label{sec:conclusion}

We presented \adapt{}, a differentiable meta-learning algorithm that adaptively allocates task proportions in multi-task instruction tuning under explicit token budgets. By updating task logits via meta-gradients of a smooth worst-case validation objective with entropy regularization, \adapt{} learns to focus training on high-utility tasks while maintaining coverage. In experiments with small open LLMs on a suite of 11 benchmarks, \adapt{} matches or slightly improves strong supervised fine-tuning baselines in accuracy and yields better training efficiency. Future work includes scaling \adapt{} to larger models and task collections, exploring alternative meta-objectives (e.g., risk-sensitive or fairness-aware), and integrating task-level allocation with example-level selection within tasks.

\bibliography{tacl2021, anthology-1, anthology-2, custom}
\bibliographystyle{acl_natbib}

\iftaclpubformat

\onecolumn

\appendix





  

\end{document}